\PassOptionsToPackage{table}{xcolor}
\documentclass[10pt,twocolumn,letterpaper]{article}
\usepackage{cvpr}              

\usepackage{colortbl}
\usepackage{multirow}
\usepackage{amsmath,amssymb}
\usepackage{graphicx}
\usepackage{caption} 

\definecolor{cvprblue}{rgb}{0.21,0.49,0.74}
\usepackage[pagebackref,breaklinks,colorlinks,allcolors=cvprblue]{hyperref}

\def\paperID{32518} 
\def\confName{CVPR}
\def\confYear{2026}

\title{RAGTrack: Language-aware RGBT Tracking with Retrieval-Augmented Generation}
\author{Hao Li$^{1,2}$, Yuhao Wang$^2$, Wenning Hao$^{1*}$, Pingping Zhang$^{2*}$, Dong Wang$^3$, Huchuan Lu$^{2,3}$\\
$^1$College of Command and Control Engineering, Army Engineering University of PLA  \\
$^2$School of Future Technology, Dalian University of Technology  \\
$^3$School of Information and Communication Engineering, Dalian University of Technology \\
{\tt\small lihao@aeu.edu.cn, 924973292@mail.dlut.edu.cn, hwnbox@aeu.edu.cn,}\\
{\tt\small\{zhpp,wdice,lhchuan\}@dlut.edu.cn}
}

\begin{document}
\maketitle
\begin{abstract}
RGB-Thermal (RGBT) tracking aims to achieve robust object localization across diverse environmental conditions by fusing visible and thermal infrared modalities. 
However, existing RGBT trackers rely solely on initial-frame visual information for target modeling,  failing to adapt to appearance variations  due to the absence of language guidance.
Furthermore, current methods suffer from redundant search regions and heterogeneous modality gaps, causing background distraction. 
To address these issues, we first introduce textual descriptions into RGBT tracking benchmarks.
This is accomplished through a pipeline that leverages Multi-modal Large Language Models (MLLMs) to automatically produce texual annotations. 
Afterwards, we propose \textbf{RAGTrack}, a novel \textbf{R}etrieval-\textbf{A}ugmented \textbf{G}eneration framework for robust RGBT tracking. 
To this end, we introduce a Multi-modal Transformer Encoder (MTE) for unified visual-language modeling. 
Then, we design an Adaptive Token Fusion (ATF) to select target-relevant tokens and perform channel exchanges based on cross-modal correlations, mitigating search redundancies and modality gaps. 
Finally, we propose a Context-aware Reasoning Module (CRM) to maintain a dynamic knowledge base and employ a Retrieval-Augmented Generation (RAG) to enable temporal linguistic reasoning for robust target modeling.
Extensive experiments on four RGBT benchmarks demonstrate that our framework achieves state-of-the-art performance across various challenging scenarios. 
The source code is available at \href{https://github.com/IdolLab/RAGTrack}{https://github.com/IdolLab/RAGTrack}.
\end{abstract}   
\renewcommand{\thefootnote}{}
\footnote{$^*$Corresponding author.}
\vspace{-3mm}
\section{Introduction}
\label{sec:intro}

Visual object tracking~\cite{song1,song2,song4,song6,qin2025must,shao2025pura,sun2024chattracker,tan2025you,tan2024xtrack,liu2025trackingmim,li2017tracking,li2025dynamic,li2024dtllm,li2023citetracker,feng2021siamese,feng2025atctrack,feng2025cstrack,gao2025tvtracker,liu2024spatial,liu2022long,zhao2022vision,zheng2025decoupled,zheng2024odtrack,zheng2022leveraging,Zheng2025umodtrack,zheng2023toward,feng2025enhancing} aims to localize objects of interest across video sequences. 
Despite its wide applications in autonomous driving, intelligent surveillance, and human-computer interaction, tracking relying solely on the visible (RGB) modality often fails under low illumination. 
In contrast, the thermal infrared (TIR) imaging has the advantage of all-weather capability, providing complementary information of the RGB modality. 
To integrate these advantages, RGB-Thermal (RGBT) tracking combines both modalities for robust performance.

\begin{figure}[t]
  \centering
  \includegraphics[width=.99\linewidth]{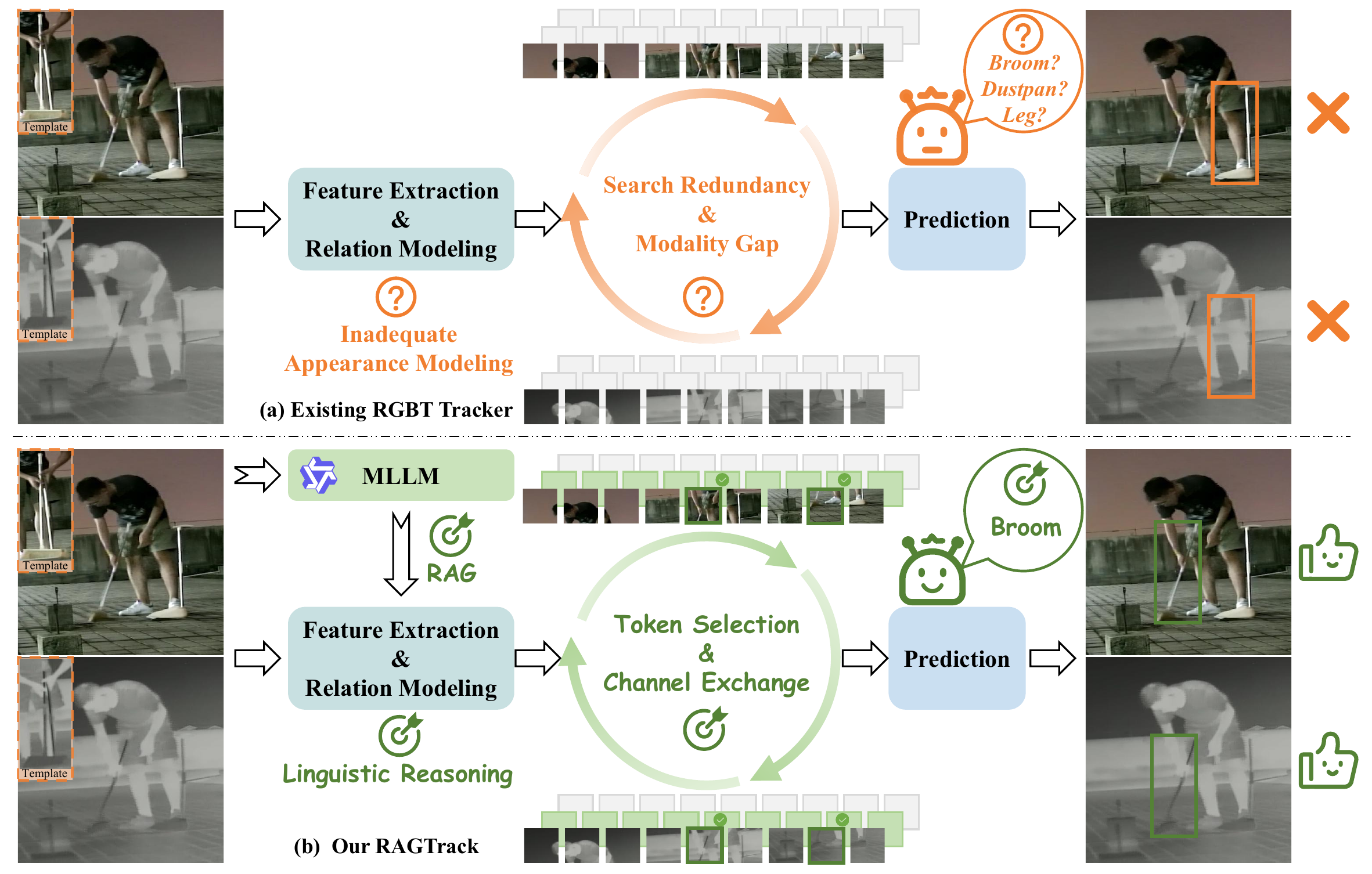}
  \vspace{-0.5em}
  \caption{Comparison with different RGBT tracking paradigms. (a) Existing RGBT trackers suffer from inadequate appearance modeling, search redundancy, and modality gap. (b) Our RAGTrack introduces linguistic reasoning, dynamic token selection, and adaptive channel exchange.}
  \label{fig:motivation}
  \vspace{-1.7em}
\end{figure}

However, current RGBT trackers~\cite{li2025cadtrack,ckd,lu2025duality,lu2025modality,lu2025after,lu2025rgbt,luo2025progressive} model the target using visual information from the initial frame, which often leads to drift under severe appearance variations. 
This limitation stems from two main factors. 
First, a single image template offers limited information of the target, failing to capture full appearance changes across different viewpoints. 
Second, the inherent ambiguity of targets often causes trackers to overemphasize unstable features, resulting in background distraction.
As shown in Fig.~\ref{fig:motivation}, the tracker may become confused among the broom, dustpan, or the lower body of the pedestrian.  
Moreover, bounding box-based initialization is inconvenient in practical scenarios, consequently hindering its widespread deployment.

On the other hand, recent advances in multi-modal object tracking have focused on integrating features from diverse modalities. 
In previous works, multi-modal features are first extracted independently, then fused through dedicated fusion modules~\cite{wang2024visual}. 
They establish connections of multiple feature sources, including RGB and TIR templates and search regions, operating at token-level.
However, a significant portion of search regions is redundant, containing substantial background regions and distractors that negatively impact the tracking accuracy. 
Furthermore, heterogeneous modality gaps hinder the establishment of effective cross-modality correspondences.
This raises a basic question: \textbf{\textit{How can we construct a target model with rich semantic representations while mitigating the impact of search redundancies and modality gaps?}}

To address these challenges, we introduce language as a high-level representation to enhance RGBT tracking. 
Language provides a more abstract understanding of objects than images, addressing semantic limitations in visual representation.
Language-based target representation can offer clear semantic information, including object categories, appearance attributes and motion states, enabling effective target-background separation.
Recent advances in RGB-Language (RGBL) tracking have demonstrated the potential of language descriptions for robust tracking~\cite{feng2025enhancing}. 
However, these methods face the challenge of visual-language misalignment as targets evolve across video frames.
Recently, Multi-modal Large Language Models (MLLMs) have exhibited strong reasoning capabilities, delivering opportunities for incorporating textual descriptions into RGBT tracking. 
Despite these advancements, there is no RGBT tracking benchmarks with text annotations. 
To bridge this gap, we extend current RGBT tracking benchmarks by introducing textual descriptions through MLLMs. 

\begin{figure*}
  \centering
  \includegraphics[width=.99\textwidth]{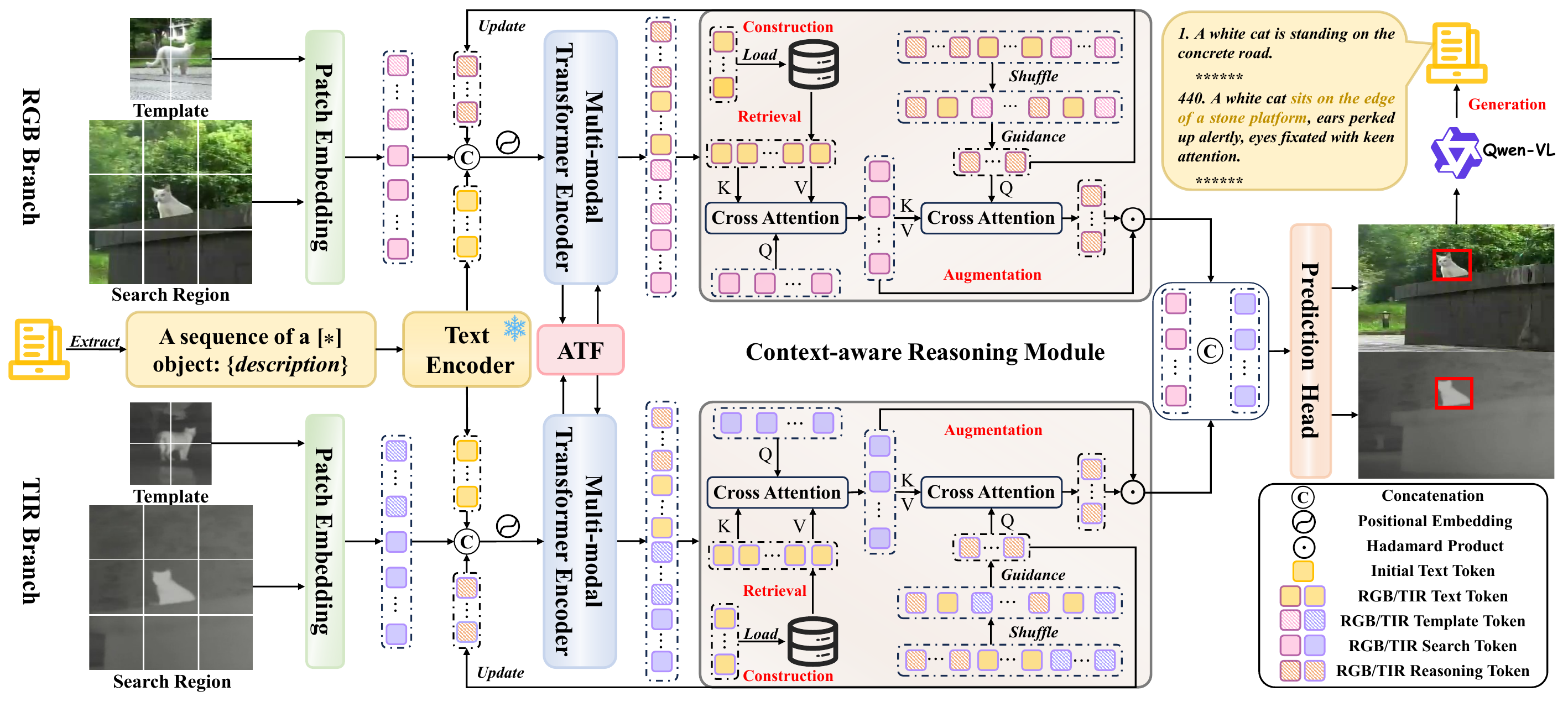}
  \vspace{-0.5em}
  \caption{Overall framework. Our method begins by tokenizing input texts and images with reasoning tokens. MTE then performs unified visual-language modeling, while ATF utilizes text-guided attention to dynamically select target-relevant tokens and enables adaptive channel exchange. Subsequently, CRM retrieves relevant contexts from a dynamic knowledge base for context-aware reasoning. Finally, the prediction head outputs tracking results, which are used by MLLMs to generate updated textual descriptions for following frames.}
  \label{fig:pipline}
  \vspace{-1.0em}
\end{figure*}

Building upon the textual descriptions, we propose a novel RGBT tracking framework, named RAGTrack. 
As depicted in Fig.~\ref{fig:pipline}, our framework consists of three key components: Multi-modal Transformer Encoder (MTE), Adaptive Token Fusion (ATF) and Context-aware Reasoning Module (CRM). 
The MTE utilizes text features for unified modeling with visual patch tokens.
Meanwhile, ATF utilizes linguistic cues to obtain text-guided attention scores for identifying target-relevant search tokens, reducing noisy interference.
This module further employs adaptive channel exchange to dynamically bridge heterogeneous features across modalities. 
As a result, it maintains semantic consistency and mitigates modality gaps.
The CRM implements a Retrieval-Augmented Generation (RAG) paradigm that establishes a continuous reasoning cycle across frames. 
This module maintains a dynamic knowledge base, retrieves historically relevant features, and generates adaptive target descriptions via MLLMs. 
With the above components, our RAGTrack enables comprehensive temporal reasoning, leveraging both visual evidence and linguistic context to maintain target identities under complex scenarios.
Extensive experiments on four benchmarks demonstrate the effectiveness of our method, achieving state-of-the-art performance.
In summary, our main contributions are as follows:
\begin{itemize}
\item To the best of our knowledge, we are the first to introduce textual descriptions into RGBT tracking, extending existing benchmarks with semantic annotations.
\item We propose RAGTrack, a novel framework that leverages Retrieval-Augmented Generation (RAG) to enhance linguistic reasoning for robust target modeling.
\item We design the Adaptive Token Fusion (ATF) to address search redundancies and modality gaps through dynamic token selection and adaptive channel exchange.
\item Extensive experiments on four RGBT tracking benchmarks demonstrate that RAGTrack outperforms existing methods in both accuracy and robustness.
\end{itemize}

\section{Related Work}
\label{sec:rel}

\subsection{RGB-Thermal Tracking}

RGBT tracking~\cite{lu2021rgbt} combines RGB and TIR modalities to overcome single-modality limitations in complex scenarios. 
Early works~\cite{zhu2019dense} use MDNet~\cite{nam2016learning} for feature extraction, while CMPP~\cite{wang2020cross} employs affinity-based pattern propagation. 
Subsequent methods advance fusion strategies, such as template-bridged interactions~\cite{tbsi} and Mamba-based fusion~\cite{lu2025rgbt}. 
The Mixture of Experts (MoE) paradigms, exemplified by XTrack~\cite{tan2024xtrack} and FlexTrack~\cite{tan2025you}, further enhance robustness through expert routing and heterogeneous fusion.
Meanwhile, temporal modeling is refined by TATrack~\cite{TATrack}, CSTrack~\cite{feng2025cstrack} and STTrack~\cite{hu2025exploiting}. 
Recently, ViPT~\cite{vipt}, BAT~\cite{BAT2024}, SDSTrack~\cite{SDSTrack}, OneTracker~\cite{OneTracker} and RDTTrack~\cite{zhu2025collaborating} leverage visual prompt learning to boost tracking performance. 
However, current methods solely depend on visual information, which becomes unstable during tracking. 
Different from existing methods, our method leverages textual descriptions to complement visual features and distinguish similar-appearance targets.

\subsection{RGB-Language Tracking}

RGBL tracking~\cite{feng2025atctrack} utilizes both natural language descriptions and visual references to localize the target across video sequences. 
Li et al.~\cite{li2017tracking} pioneer this field by extending tracking datasets with linguistic annotations. 
Wang et al.~\cite{wang2021towards} and Zhou et al.~\cite{zhou2023joint} advance visual grounding techniques to integrate language semantics with tracking. 
Subsequently, SNLT~\cite{feng2021siamese}, All-in-One~\cite{zhang2023all} and OVLM~\cite{zhang2023one} develop multi-modal fusion approaches through dynamic aggregation, unified Transformers and memory mechanism, respectively. 
CiteTracker~\cite{li2023citetracker} and MMTrack~\cite{zheng2023toward} adopt generative frameworks for text-based target estimation, while VLTTT~\cite{guo2024divert} and UVLTrack~\cite{ma2024unifying} leverage contrastive learning for vision-language alignment. 
Recently, LLM-driven methods, such as DTLLM-VLT~\cite{li2024dtllm}, ChatTracker~\cite{sun2024chattracker}, DUTrack~\cite{li2025dynamic} and ReasoningTrack~\cite{wang2025reasoningtrack}, enhance robustness through multimodal reasoning and dynamic language generation. 
However, current methods suffer from insufficient visual-language feature association and ineffective background interference suppression. 
Different from existing methods, our method utilizes text features to address search redundancies and modality gaps through dynamic token selection and adaptive channel exchange.

\subsection{Retrieval-Augmented Generation}

RAG~\cite{lewis2020retrieval} can enhance LLMs by retrieving external knowledge during inference to mitigate hallucinations and improve the factual accuracy in domain-specific tasks. 
While widely adopted in applications~\cite{guo2025deepseek,Tang2025MBARAG}, its integration into visual tracking remains limited. 
As an outstanding work, TrackingMiM~\cite{liu2025trackingmim} proposes a query-based retrieval module for UAV tracking.
However, current methods merely reuse pre-stored features.
Different from existing methods, our method introduces RAG into RGBT tracking for the first time, enabling context-aware linguistic reasoning.

\section{Methodology}

In this paper, we propose RAGTrack, which includes three key components: Multi-modal Transformer Encoder (MTE), Adaptive Token Fusion (ATF), and Context-aware Reasoning Module (CRM). 
As shown in Fig.~\ref{fig:pipline},
MTE serves as the foundation for processing heterogeneous inputs. 
ATF dynamically captures target-relevant tokens and fuses cross-modal features. 
CRM leverages a RAG mechanism to maintain target identities through temporal semantic reasoning.
Details are described as follows.

\subsection{Overall Framework}

Our framework localizes targets across video sequences by fusing complementary information from RGB, TIR, and language modalities. 
At time step $t$, the input consists of search images $\mathbf{X}_m^t \in \mathbb{R}^{3 \times H_x \times W_x}$ and multi-modal references comprising template images $\mathbf{Z}_m^t \in \mathbb{R}^{3 \times H_z \times W_z}$ and language descriptions \(\mathbf{L}^t\). 
Here, \(H\) and \(W\)  denote the height and width of the images, and $m \in \{B,R\}$ represents the RGB or TIR modality. 
The MTE employs parameter-shared branches for processing both RGB and TIR modalities. 
The encoder processes these inputs via patch embedding and tokenization to generate tokens, which are concatenated and fed into the visual-language unified modeling. 
These tokens then undergo ATF to mitigate search redundancies and modality gaps, followed by CRM for linguistic refinement. 
Finally, the refined features are fed into the prediction head to produce tracking results.

\subsection{Multi-modal Transformer Encoder}

To extract multi-modal features, we introduce MTE for unified visual-language modeling.
We process template and search images through a three-stage downsampling~\cite{zhang2023hivit}.
It transforms the template and search images into patch tokens $\hat{\mathbf{Z}}_m^t \in \mathbb{R}^{N_z \times C}$ and $\hat{\mathbf{X}}_m^t \in \mathbb{R}^{N_x \times C}$, where \(N\) represents the number of tokens, and  \(C\) represents the feature dimension.

Afterwards, we introduce a sequence prefix $\mathbf{E}^t$ to enhance the temporal awareness and address the potential misalignment between visual contents and language descriptions across frames. 
This prefix combines a fixed textual prompt with learnable tokens. 
Specifically, $\mathbf{E}^t$ is formulated as \textit{“A sequence of a [$ * $] object:”}, where \textit{[$ * $]} is replaced by learnable tokens.
We concatenate $\mathbf{E}^t$ and $\mathbf{L}^t$ to form the text input $\mathbf{H}^t = [\mathbf{E}^t, \mathbf{L}^t]$.
Then, $\mathbf{H}^t$ is encoded by the text encoder $\mathcal{T}$ to obtain semantic features $\hat{\mathbf{H}}^t\in \mathbb{R}^{N_h \times C}$:
\begin{equation}
    \hat{\mathbf{H}}^t = \mathcal{T}(\mathbf{H}^t).
\end{equation}

Subsequently, we concatenate the reasoning token $\mathbf{R}_{m}^{t}\in \mathbb{R}^{N_r \times C}$, the text token $\hat{\mathbf{H}}^t$, the template token $\hat{\mathbf{Z}}_m^t$ and the search token $\hat{\mathbf{X}}_m^t$ into a unified token sequence:
\begin{equation}
\mathbf{F}_m^0= \left[ \mathbf{R}_{m}^{t}; \hat{\mathbf{H}}^t;\hat{\mathbf{Z}}_m^t; \hat{\mathbf{X}}_m^t \right],
\end{equation}
where \([;]\) denotes the concatenation operation. The unified visual-language modeling is formulated as:
\begin{equation}
\begin{aligned}
\mathbf{\hat{F}}_m^{l-1}&={\rm{MHSA}}(\mathbf{F}_m^{l-1},\mathbf{F}_m^{l-1},\mathbf{F}_m^{l-1}), \\
\tilde{\mathbf{F}}_m^{l-1}&=\mathbf{F}_m^{l-1}+{\rm{LN}}(\delta_{1}\cdot \mathbf{\hat{F}}_m^{l-1}), \\
\mathbf{F}_m^{l}&=\tilde{\mathbf{F}}_m^{l-1}+{\rm{LN}}(\delta_{2}\cdot {\rm{MLP}}(\tilde{\mathbf{F}}_m^{l-1})),
\end{aligned}
\end{equation}
where \(\rm{MHSA}\) stands for multi-head self-attention~\cite{vit}, \(\rm{LN}\) represents the layer normalization~\cite{ba2016layer} and \(\rm{MLP}\) denotes the multilayer perceptron. 
$l \in \left\{ {1,2,\dots,L} \right\}$ indicates the layer index.
\(\delta_{1}\) and \(\delta_{2}\) are two learnable parameters.
By unified visual-language modeling, semantic information from text is effectively leveraged to enhance feature discrimination.

\subsection{Adaptive Token Fusion}

\begin{figure}[t]
  \centering
  \includegraphics[width=0.96\linewidth]{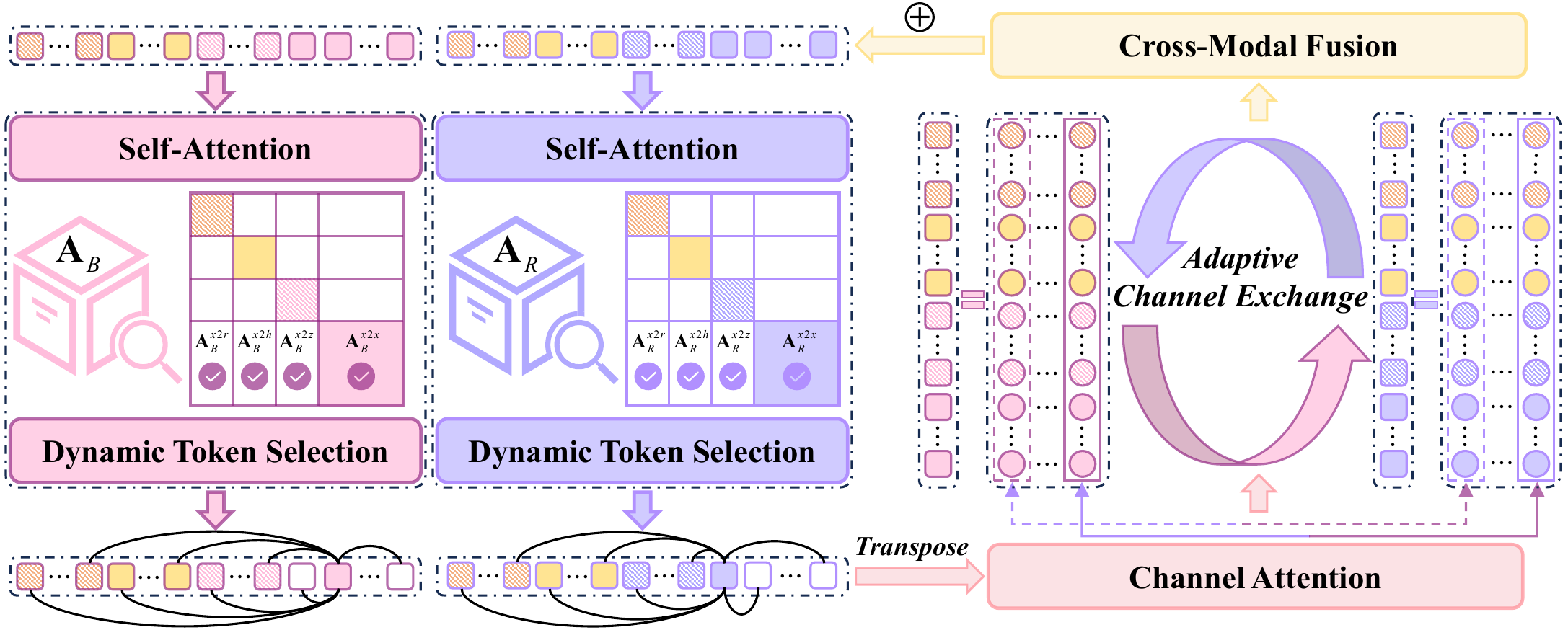}
  \vspace{-0.5em}
  \caption{Details of our proposed ATF.}
  \label{fig:ATCF}
  \vspace{-1.5em}
\end{figure}

To address search redundancies and modality gaps, we propose ATF for dynamic token selection and adaptive channel exchange. 
Traditional fusion methods suffer from inefficient token utilization and significant feature discrepancies, leading to degraded tracking performance. 
In contrast, ATF identifies target-relevant tokens and adapts channel exchanges based on cross-modality correlations.

In fact, attention scores $\mathbf{A}_m$ from the self-attention module serve as an effective indicator for token significance, while simultaneously guiding the model to excavate contextual information. 
This fact motivates our use of attention scores as metrics for token selection. 
As shown in Fig.~\ref{fig:ATCF}, the calculation for each term is as follows:
\begin{equation}
\begin{aligned}
&\mathbf{A}_m^{x2o} = \text{Softmax}\left(\frac{\mathbf{Q}_m^x(\mathbf{K}_m^o)^{\rm T}}{\sqrt{d}}\right),
\end{aligned}
\end{equation}
where $\mathbf{A}_m^{x2r}$, $\mathbf{A}_m^{x2h}$, $\mathbf{A}_m^{x2z}$, and $\mathbf{A}_m^{x2x}$ represent correlations between search and reasoning, text, template, and search tokens, respectively. 
Specially, to mitigate template noise, we extract the center region of the template image, as it contains sufficient target information~\cite{ostrack}. 
We aggregate these scores to identify key tokens as follows:
\begin{equation}
\begin{aligned}
\label{eq:attn_sum}
\mathbf{A}_m^{\text{total}} = \mathbf{A}_m^{x2r} + \mathbf{A}_m^{x2h} + \mathbf{A}_m^{x2z} + \mathbf{A}_m^{x2x}.
\end{aligned}\end{equation}
Tokens with higher $\mathbf{A}_m^{\text{total}}$ are retained using a retention ratio $\gamma$. 
This parameter-free token selection avoids redundant computations by directly reusing the attention scores.

For channel exchange, we address the challenge of modality gaps by establishing channel-level associations between modalities.
More specifically, we first compute cross-modal relevance $\mathbf{S}$ between features along the channel dimension:
\begin{equation}
\mathbf{S} = \left(\left(\mathbf{F}_B^l\right)^{\rm T} \mathbf{W}_B^l\right) \left(\left(\mathbf{F}_R^l\right)^{\rm T} \mathbf{W}_R^l\right)^{\rm T},
\end{equation}
where $\mathbf{W}_B^l$ and $\mathbf{W}_R^l$ are learnable parameters.
Then, we average $\mathbf{S}$ along the channel dimension to determine channel importance, then select corresponding channels for adaptive exchange based on the exchange ratio $\sigma $. 
Finally, we concatenate the features of RGB and TIR modalities along the token dimension and apply the cross-modal fusion $\mathcal{M}$:
\begin{equation}
\mathbf{F}_B^l, \mathbf{F}_R^l = \mathcal{M}\left(\left[\mathbf{F}_B^l; \mathbf{F}_R^l\right]\right),
\end{equation}
where $\mathcal{M}$ is a MLP layer.
This channel exchange bridges the independently processed RGB and TIR modalities during feature extraction, enabling the model to learn cross-modal correlations and enhance feature discrimination.

Our ATF mitigates search redundancies and modality gaps through dynamic token selection and adaptive channel exchange. 
It facilitates information transmission between RGB and TIR modalities,  enabling the model to focus on the most relevant features for enhanced robustness.

\subsection{Context-aware Reasoning Module}

To enhance temporal modeling, we propose the CRM for linguistic reasoning. 
As illustrated in Fig.~\ref{fig:pipline}, it includes a RAG mechanism that operates through four stages: construction, retrieval, augmentation, and generation.

\textbf{Construction.} 
We construct a local knowledge base from historical text features. 
This knowledge base is maintained as a set of $n$ feature embeddings ${\mathbf{D}_m} = \left\{ {\hat{\mathbf{H}}_m^1, \cdots ,\hat{\mathbf{H}}_m^n} \right\}$, which is dynamically updated during inference to suppress redundancy.
Specifically, a new text feature $\hat{\mathbf{H}}_m^t$ is added only if its maximum cosine similarity with existing entries falls below a threshold $\lambda$:
\begin{equation}
\mathop {\max }\limits_{\hat{\mathbf{H}}_m^i \in {\mathbf{D}_m}} \frac{{\hat{\mathbf{H}}_m^t\cdot\hat{\mathbf{H}}_m^i}}{{{{\left\| {\hat{\mathbf{H}}_m^t} \right\|}_2}{{\left\| {\hat{\mathbf{H}}_m^i} \right\|}_2}}} < \lambda.
\end{equation}

\textbf{Retrieval.} 
Given a query feature $\hat{\mathbf{H}}_m^t$, the retriever $\mathcal{O}$ selects the top-$k$ most relevant features from $\mathbf{D}_m$:
\begin{equation}
{\mathbf{V}_m} = \mathcal{O}\left( {\hat{\mathbf{H}}_m^t,{\mathbf{D}_m}} \right),
\end{equation}
where ${\mathbf{V}_m} \subseteq {\mathbf{D}_m}$ and $\left| {{\mathbf{V}_m}} \right| = k$. 
To refine the search features, we then apply intra-modal cross-attention $\Phi$:
\begin{equation}
\bar{\mathbf{X}}_m^t = \hat{\mathbf{X}}_m^t + \Phi\left( \hat{\mathbf{X}}_m^t, {\mathbf{V}_m} \right).
\end{equation}

\textbf{Augmentation.} 
For temporal propagation across video frames, we enhance the reasoning tokens to capture both local and global context. 
First, we perform an average pooling $\mathcal{P}$ on the reasoning features, text features, and template features along the token dimension:
\begin{equation}
\bar{\mathbf{R}}_m^{t},\bar{\mathbf{H}}_m^t,\bar{\mathbf{Z}}_m^t = {\mathcal{P}}({\mathbf{R}}_m^{t},\hat{\mathbf{H}}_m^t,\hat{\mathbf{Z}}_m^t).
\end{equation}

Subsequently, we concatenate these pooled features along the channel dimension and apply the guidance $\mathcal{G}$ to inject current frame cues:
\begin{equation}
{\mathbf{R}}_m^{t+1} = \mathcal{G}(\bar{\mathbf{R}}_m^{t},\bar{\mathbf{H}}_m^t,\bar{\mathbf{Z}}_m^t),
\end{equation}
where $\mathcal{G}$ is a MLP layer. 
Then, the updated reasoning token ${\mathbf{R}}_m^{t+1}$ is propagated to the next frame. 

Finally, we enhance the feature representation through a three-step temporal augmentation:
\begin{equation}
\begin{aligned}
\mathbf{\hat{R}}_m^{t+1} &= \mathbf{R}_m^{t+1} + \Phi(\mathbf{R}_m^{t+1}, \bar{\mathbf{X}}_m^t), \\
\mathbf{\tilde{R}}_m^{t+1} &= \mathbf{\hat{R}}_m^{t+1} + \text{MLP}(\mathbf{\hat{R}}_m^{t+1}), \\
\tilde{\mathbf{X}}_m^t &= \bar{\mathbf{X}}_m^t \otimes (\mathbf{\tilde{R}}_m^{t+1})^{\rm T} \odot \bar{\mathbf{X}}_m^t,
\end{aligned}
\end{equation}
where $\otimes$ and $\odot$ denotes the matrix multiplication and hadamard product. 

\textbf{Generation.} 
To overcome the limitations of static language annotations, we employ MLLMs to dynamically generate context-aware target descriptions during inference. 
The input consists of the search image and a structured prompt: \textit{“Describe the object located in the image at $<$box$>$({x},{y},{x+w},{y+h})$<$/box$>$. Focus on distinctive visual features, motion patterns, and key identifiers to distinguish it from background elements and distractors.”} 
The resulting descriptions continuously refresh the multi-modal reference, improving cross-frame appearance reasoning.

Our CRM maintains a dynamic knowledge base through iterative retrieval and update mechanisms, and enhances temporal coherence via reasoning token propagation. 
It further generates adaptive target descriptions to overcome static annotation limitations and adapt to scene variations.

\begin{table*}[t]
    \centering
     \caption{Performance comparison on four RGBT tracking benchmarks. The best results are in bold.}
     \vspace{-0.9em}
    \setlength{\tabcolsep}{1.4mm}{
    \renewcommand\arraystretch{0.89}{
	\begin{tabular}{c|c|c|cc|cc|cc|ccc}
        \toprule
		\multirow{2}{*}{Method} & \multirow{2}{*}{Publication} &\multirow{2}{*}{Resolution} & \multicolumn{2}{c|}{GTOT} & \multicolumn{2}{c|}{RGBT210} & \multicolumn{2}{c|}{RGBT234} & \multicolumn{3}{c}{LasHeR} \\\cline{4-12}
		 & & & \rule{0pt}{2.2ex}MPR$\uparrow$ & MSR$\uparrow$ & PR$\uparrow$ & SR$\uparrow$ & MPR$\uparrow$ & MSR$\uparrow$ & PR$\uparrow$  & NPR$\uparrow$  & SR$\uparrow$  \\
		\midrule
               
    CMPP~\cite{wang2020cross} & CVPR 2020 & 256$\times$256 &  92.6 & 73.8 & $-$ & $-$ & 82.3& 57.5 & $-$ & $-$  & $-$  \\
    CAT~\cite{li2020challenge} & ECCV 2020 & 256$\times$256 & 88.9 & 71.7 & 79.2  & 53.3 & 80.4& 56.1 & 45.0  & 39.5  & 31.4  \\
    APFNet~\cite{xiao2022attribute} & AAAI 2022 & 256$\times$256 & 90.5 & 73.7 & $-$  & $-$ & 82.7& 57.9 & 50.0  & 43.9  & 36.2  \\
    CMD~\cite{zhang2023efficient} & CVPR 2023 &  256$\times$256 & 89.2 & 73.4 & $-$ & $-$ & 82.4& 58.4 & 59.0  &54.6 & 46.4  \\
    ViPT~\cite{vipt} & CVPR 2023 &  256$\times$256 & $-$ & $-$ & $-$ & $-$ & 83.5 & 61.7 & 65.1 & $-$ & 52.5 \\
    TBSI~\cite{tbsi} & CVPR 2023 &  256$\times$256 & $-$ & $-$ & 85.3 & 62.5 & 87.1 & 63.7 & 69.2 & 65.7 & 55.6  \\
    TATrack~\cite{TATrack} & AAAI 2024  & 256$\times$256 & $-$ & $-$ & 85.3 & 61.8 & 87.2 & 64.4 & 70.2 & 66.7 & 56.1  \\
    BAT~\cite{BAT2024} & AAAI 2024  & 256$\times$256 & $-$ & $-$ & $-$ & $-$ & 86.8 & 64.1 & 70.2& $-$ & 56.3  \\
    GMMT~\cite{gmmt} & AAAI 2024  & 256$\times$256 & $-$ & $-$ & $-$ & $-$ & 87.9 & 64.7 & 70.7 & 67.0 & 56.6  \\
    {Un-Track}~\cite{Un-Track} & CVPR 2024  & 256$\times$256 & $-$ & $-$ & $-$ & $-$ & 83.7 & 61.8 & 66.7& $-$ & 53.6  \\
    SDSTrack~\cite{SDSTrack} & CVPR 2024  & 256$\times$256 & $-$ & $-$ & $-$ & $-$ & 84.8 & 62.5 & 66.5& $-$ & 53.1 \\
    OneTracker~\cite{OneTracker} & CVPR 2024  & 384$\times$384 & $-$ & $-$ & $-$ & $-$ & 85.7 & {64.2} & 67.2& $-$ & 53.8  \\
    CKD~\cite{ckd} & ACM MM 2024 & 256$\times$256 & 93.2 & 77.2 & 88.4 & 65.2 & 90.0 & 67.4 & 73.2& 69.3 & 58.1  \\
    PTrMA~\cite{luo2025progressive} &  TIM 2025 & 256$\times$256 & $-$ & $-$ & 86.0 & 63.4 & 87.6 & 65.2 & 71.5 & 67.5 & 56.8  \\ 
    TVTracker~\cite{gao2025tvtracker}  &  IoTJ 2025 & 256$\times$256 & $-$ & $-$ & 87.6 & 63.7 & 88.6 & 64.8 & 72.6 & 68.4 & 57.5  \\ 
    AINet~\cite{lu2025rgbt}  &  AAAI 2025 & 384$\times$384 & $-$ & $-$ & 87.5 & 64.8 & 89.2 & 67.3 & 74.2 & 70.1 & 59.1  \\
    SUTrack~\cite{chen2025sutrack} &  AAAI 2025 & 384$\times$384 & $-$ & $-$ &$-$ & $-$ & 92.1 & 69.2 & 75.8& $-$ & 60.9 \\
    STTrack~\cite{hu2025exploiting} &  AAAI 2025 & 256$\times$256 & $-$ & $-$ &$-$ & $-$ & 89.8 & 66.7 & 76.0 & $-$ & 60.3 \\
    CAFormer~\cite{xiao2025cross}  &  AAAI 2025 & 256$\times$256 & 91.8 & 76.9 & 85.6 & 63.2 & 88.3 & 66.4 & 70.0 &66.1& 55.6 \\
    TBSI-Ext~\cite{li2024rgb} & TPAMI 2025 & 256$\times$256 & $-$ & $-$ & $-$ & $-$ & 91.0 & 67.0 & 75.5& 71.5  & 59.6 \\
    CMDTrack~\cite{zhang2025cross} & TPAMI 2025 & 256$\times$256 & $-$ & $-$ & $-$ & $-$ & 85.9 & 61.8 & 68.8& $-$  & 56.6 \\
    UM-ODTrack~\cite{Zheng2025umodtrack} &  TPAMI 2025 &  384$\times$384 & $-$ & $-$ & $-$ & $-$ & 91.5 & 69.2 & 74.4 & $-$ & 58.8 \\
    XTrack~\cite{tan2024xtrack}  &  ICCV 2025 & 256$\times$256 & $-$ & $-$ & $-$ & $-$ & 87.4 & 64.9 & 69.1 & $-$ & 55.7 \\
    SMSTracker~\cite{chan2025smstracker} &  ICCV 2025 &  256$\times$256 & $-$ & $-$ & $-$ & $-$ & 86.9 & 64.5 & 70.3 & $-$ & 56.0 \\
    AETrack~\cite{zhu2025adaptive} &  TCSVT 2025 & 256$\times$256 & $-$ & $-$ & 90.4 & 66.3 & 91.6 & 68.8 & 74.7& 71.0 & 59.6  \\
    MambaVT~\cite{lai2025mambavt}  &  TCSVT 2025  & 256$\times$256 & 94.1 & 75.3 & 88.0 & 63.7 & 88.9 & 65.8 & 73.0 & 69.5 & 57.9  \\
    MoETrack~\cite{tang2025revisiting}  &  TIP 2025 & 256$\times$256 & 93.6 & 78.4 & $-$ & $-$ & 88.1 & 65.1 & 72.1 & $-$ & 57.8 \\
    DMD~\cite{hu2025dual}  &  TIP 2025 & 256$\times$256 & 94.2 & 78.6 & 89.6 & 65.8 & 90.7 & 67.8 & 73.6 &69.7 & 58.2 \\ 
    \hline
    \rowcolor[gray]{0.92}
    \rule{0pt}{2.2ex}RAGTrack &  Ours  & 256$\times$256 & \textbf{95.1} & \textbf{79.3} & \textbf{93.2} & \textbf{67.1} & \textbf{93.8} & \textbf{69.5} & \textbf{76.8} & \textbf{73.0} & \textbf{61.1} \\  
    \noalign{\hrule height 0.8pt}
       \end{tabular}}}

\label{overall_result}
\vspace{-0.7em}
\end{table*}
\subsection{Prediction Head and Loss Function}

The refined multi-modal features $\tilde{\mathbf{X}}_m^t$ are first reshaped into a 2D spatial representation. 
These features are then concatenated, followed by channel compression with convolutional layers.
Finally, they are transformed through a Fully Convolutional Network (FCN) consisting of stacked Conv-BN-ReLU layers~\cite{nair2010rectified,ioffe2015batch}.
The prediction head generates three outputs: classification scores, spatial offsets, and normalized sizes.
The final bounding box is constructed at the position with maximum classification score by combining the corresponding predictions.

For model optimization, we adopt a multi-task loss function combining the focal loss $L_{\text{cls}}$~\cite{law2018cornernet} for classification, along with $L_1$ loss and generalized IoU loss $L_{\text{iou}}$~\cite{rezatofighi2019generalized} for regression. 
The overall loss function is formulated as:
\begin{equation}
\mathcal{L} = L_{\text{cls}} + \lambda_{\text{iou}} L_{\text{iou}} + \lambda_{L_1} L_1,
\end{equation}
where \(\lambda_{\text{iou}}\) and \(\lambda_{L_1}\) are the hyper-parameters.

\section{Experiment}
\subsection{Datasets and Evaluation Metrics}

Our experiments are conducted on four RGBT tracking benchmarks. 
GTOT~\cite{li2016gtot} contains 50 RGB-T sequences covering seven primary challenges. 
RGBT210~\cite{Li17rgbt210} contains 210 sequences and 12 challenging attributes.
RGBT234~\cite{li2019rgb234} includes 234 sequences with increased diversities. 
LasHeR~\cite{li2021lasher} comprises 1,224 sequences annotated with 19 fine-grained attributes.
To enable language-aware RGBT tracking, we extend these benchmarks with textual descriptions through a two-step generation pipeline.
In the first step, we generate informative text descriptions by inputting the image and bounding box into MLLMs, overcoming the time-consuming and labor-intensive nature of manual annotations.
Then, we refine these generated descriptions through MLLMs and human experts to mitigate hallucinations, ensuring high-quality semantic annotations.
For the training set of LasHeR, we annotate all frames across 979 sequences with 514,081 textual descriptions to support model training. 
The test sets of LasHeR and the other three benchmarks are annotated with first-frame descriptions to evaluate the textual reasoning capability during inference.
Following standard protocols, we adopt the Precision Rate (PR) and Success Rate (SR) as primary evaluation metrics. 
For LasHeR, we additionally use the Normalized Precision Rate (NPR) to account for target scale variations. 
On GTOT and RGBT234 with modality misalignment, we report the Maximum Precision Rate (MPR) and Maximum Success Rate (MSR) for fair comparison.
\subsection{Implementation Details}
\begin{figure}[t]
  \centering
  \includegraphics[width=1.0\linewidth]{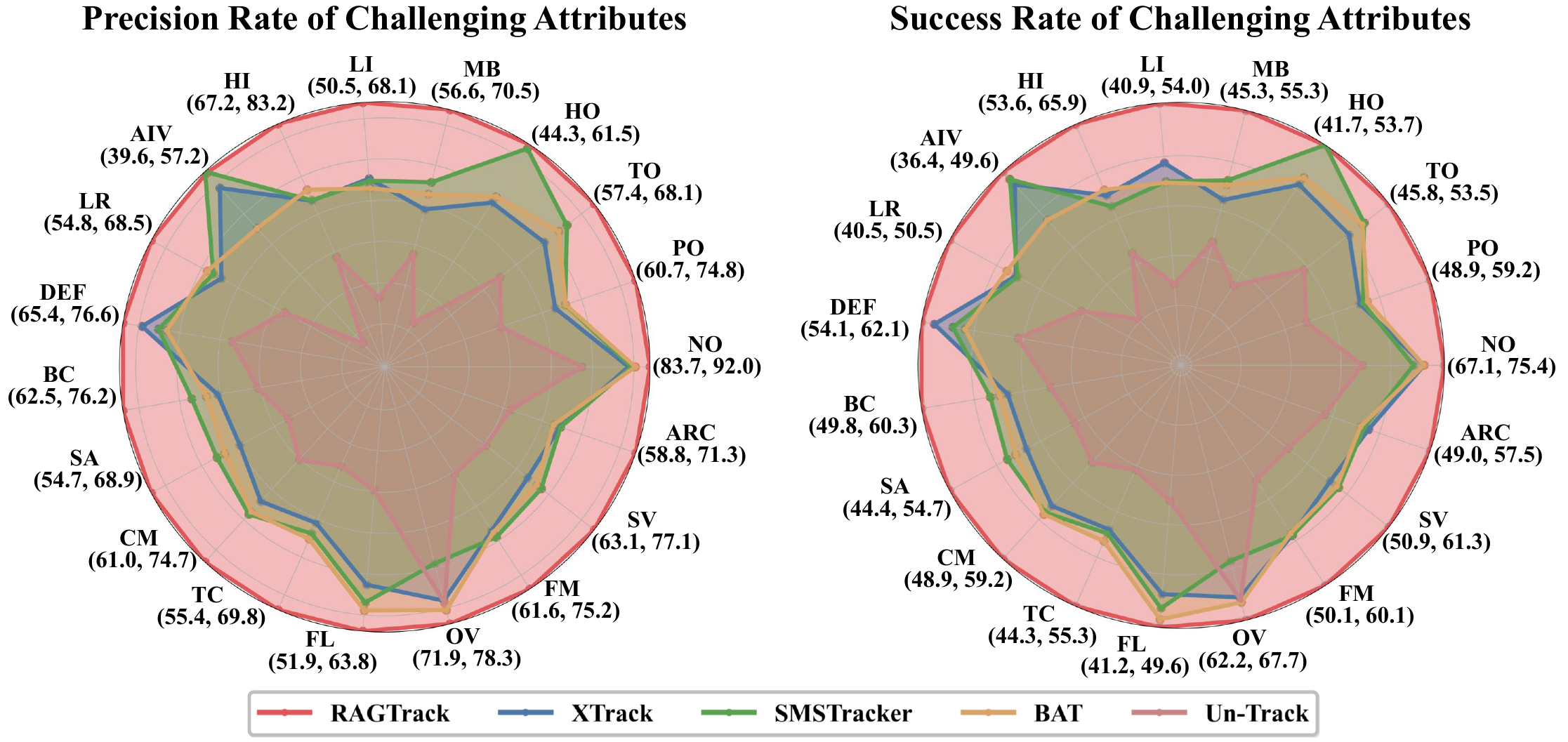}
  \vspace{-6mm}
  \caption{Attribute-based evaluations on the LasHeR dataset.}
  \label{fig:radar_plot}
  \vspace{-6mm}
\end{figure}
We implement RAGTrack using PyTorch on 4 NVIDIA V100 GPUs with a batch size of 16. 
The text encoder employs CLIP~\cite{radford2021learning}, while the visual backbone uses HiViT-B initialized from SOT~\cite{li2025dynamic}. 
Training employs the AdamW optimizer~\cite{AdamW} with a learning rate $10^{-4}$ and a weight decay $10^{-4}$.
Template images and search regions are processed at fixed resolutions of 128$\times$128 and 256$\times$256, respectively.
We configure feature dimension $C=512$, text token $N_h=1$, and reasoning token $N_r=1$.
Our ATF is deployed at the 6-th, 12-th, 18-th and 24-th layers of HiViT-B with retention ratio $\gamma=85\%$.
We select 256 channels for exchange per modality.
In CRM, the local knowledge base maintains $n = 4$ entries, and the retrieval selects $k = 2$ features with threshold $\lambda=1.0$. 
The MLLM adopts QWen2.5-VL-3B~\cite{Qwen2VL}.
We train RAGTrack on the LasHeR training set with loss weights \(\lambda_{\text{iou}}=2\) and \(\lambda_{L_1}=5\) to balance the regression and classification terms.
\subsection{Comparison with State-of-the-Art Trackers}
We evaluate RAGTrack on four RGBT benchmarks against state-of-the-art methods. 
Quantitative results are summarized in Tab.~\ref{overall_result}, with detailed analysis provided below.

\textbf{GTOT.} 
Our method delivers state-of-the-art performance with MPR 95.1\% in Tab.~\ref{overall_result}. 
These results demonstrate significant improvements: +1.5\% MPR over MoETrack~\cite{tang2025revisiting} and +4.0\% MSR over MambaVT~\cite{lai2025mambavt}. 
The performance gain highlights the effectiveness of our method in handling complex scenarios with modality discrepancies.

\textbf{RGBT210.} 
Our method shows outstanding performance, achieving 93.2\% PR and 67.1\% SR in Tab.~\ref{overall_result}. 
It exceeds AETrack~\cite{zhu2025adaptive} by a margin of +2.8\% PR and shows improved robustness than AINet~\cite{lu2025rgbt} with a margin of +2.3\% SR. 
These results indicate the robustness of our language-aware representation in improving tracking accuracy across diverse environmental conditions.

\textbf{RGBT234.} 
Our method achieves 93.8\% MPR and 69.5\% MSR in Tab.~\ref{overall_result}, outperforming SMSTracker~\cite{chan2025smstracker} by 6.9\% MPR and STTrack~\cite{hu2025exploiting} by 2.8\% MSR. 
This improvement validates the superiority of our RAGTrack in leveraging semantic information for accurate target localization.

\textbf{LasHeR.} 
Our method achieves leading performance with 76.8\% PR and 61.1\% SR in Tab.~\ref{overall_result}, surpassing recent trackers: +4.2\% PR over TVTracker~\cite{gao2025tvtracker} and +5.4\% SR over XTrack~\cite{tan2024xtrack}.  
Fig.~\ref{fig:radar_plot} further reveals the attribute-specific strengths: +10.7\% PR in Total Occlusion (TO) and +5.5\% SR in Out-of-View (OV). 
Notably, these attribute-specific advantages underscore the ability of our CRM in maintaining target identities under appearance variations.
\subsection{Ablation Studies}
We conduct thorough ablation studies to evaluate key components and parameters in RAGTrack, with results summarized in the corresponding tables and figures below.
\begin{table}[t]
\centering
\caption{Ablation study of key components on RGBT234.}
\label{tab:ablation}
\vspace{-0.9em}
\renewcommand\arraystretch{0.89}{
\begin{tabular}{ccc}
\toprule
Method  & MPR$\uparrow$  & MSR$\uparrow$ \\
\midrule
Baseline  & 87.9  & 64.5 \\
w/ CRM*  &  89.1 & 65.0 \\
w/ MTE+CRM*  &  91.1 & 66.7 \\
w/ MTE+CRM  &  91.8 & 67.4 \\
\rowcolor[gray]{0.92}
\rule{0pt}{2.0ex}w/ MTE+CRM+ATF  & \textbf{93.8}  & \textbf{69.5} \\
\noalign{\hrule height 0.8pt}
\end{tabular}
}
\vspace{-1.1em}
\end{table}
\begin{table}[t]
\centering
\caption{Comparison of fusion positions on RGBT234.}
\label{tab:position}
\vspace{-0.9em}
\renewcommand\arraystretch{0.89}{
\begin{tabular}{cccccc}
\toprule
 6 & 12 & 18 & 24 & MPR$\uparrow$  & MSR$\uparrow$ \\
\midrule
  \checkmark & & & &  92.2  &  67.9  \\
  \checkmark & \checkmark & & &  92.5   & 68.5  \\
    \checkmark & \checkmark & \checkmark & &  92.9   &  68.8 \\
  \rowcolor[gray]{0.92}
  \rule{0pt}{2.0ex}\checkmark & \checkmark & \checkmark & \checkmark & \textbf{93.8} & \textbf{69.5} \\
  \noalign{\hrule height 0.8pt}
\end{tabular}
}
\vspace{-1.1em}
\end{table}
\begin{table}[t]
\centering
\caption{Comparison of fusion paradigms on RGBT234.}
\label{tab:fusion_type}
\vspace{-0.9em}
\renewcommand\arraystretch{0.89}{
\begin{tabular}{ccccc}
\toprule
Module & MPR$\uparrow$  & MSR$\uparrow$ & Params$ \downarrow $  \\
\midrule
TBSI~\cite{tbsi} &  92.8 & 67.6 & 145.9M  \\
BSI~\cite{hu2025exploiting} & 93.1   & 68.2 & 103.6M  \\
DFM~\cite{lu2025rgbt}  &  92.7 & 67.8 & 110.3M   \\
\rowcolor[gray]{0.92}
\rule{0pt}{2.0ex}ATF (Ours) & \textbf{93.8}  & \textbf{69.5} & \textbf{101.8M} \\
\noalign{\hrule height 0.8pt}
\end{tabular}
}
\vspace{-1.4em}
\end{table}

\textbf{Component Analysis.}
We conduct ablation studies to evaluate the contribution of key components.
As detailed in Tab.~\ref{tab:ablation}, the baseline with a backbone and convolutional fusion achieves 87.9\% MPR.
The text-free CRM* brings moderate improvement to 89.1\% MPR, while incorporating MTE with CRM* further increases performance to 91.1\% MPR. 
The complete CRM yields more clear gains at 91.8\% MPR, confirming the value of language information.
The full integration of MTE, CRM and ATF achieves superior performance at 93.8\% MPR, validating their complementary roles in semantic reasoning and feature interaction.
These results indicate that both linguistic enhancement and cross-modal fusion are essential for robust RGBT tracking.

\textbf{Effect of Fusion Positions.} 
As shown in Tab.~\ref{tab:position}, fusions at Layers 6, 12, 18 and 24 achieve the optimal performance. 
Early fusions capture low-level visual features but lack semantics, while deeper fusions extract high-level contextual information but miss fine-grained details. 
The progressive fusion achieves complementary advantages, showing the importance of cross-layer semantic fusion.

\textbf{Effect of Different Fusion Paradigms.} 
To validate the effectiveness of ATF, we compare it with several representative fusion paradigms: 
TBSI~\cite{tbsi} establishes template-bridged interaction, 
BSI~\cite{hu2025exploiting} introduces template-temporal similarity for background suppression
and DFM~\cite{lu2025rgbt} models inter-modal differences for complementary fusion.
In contrast, ATF selects target-relevant tokens and exchanges informative channels. 
As shown in Tab.~\ref{tab:fusion_type}, ATF achieves the best performance (69.5\% MSR) with the fewest trainable parameters (101.8M). 
The results confirm that ATF enables more discriminative and efficient cross-modal fusion.

\textbf{Effect of Different Augmentation Mechanisms.} 
We evaluate different designs for the augmentation in CRM. 
As shown in Tab.~\ref{tab:Augmentation_type}, our method achieves superior performance. 
The addition operation yields a limited accuracy due to insufficient representational capacity. 
Transformer increases trainable parameters by 3.0\%, whereas Mamba shows lower performance.
These results confirm that our design provides the balance for temporal propagation.
\begin{table}[t]
\centering
\caption{Comparison of augmentation mechanisms on RGBT234.}
\label{tab:Augmentation_type}
\vspace{-0.9em}
\renewcommand\arraystretch{0.89}{
\begin{tabular}{ccccc}
\toprule
Module & MPR$\uparrow$  & MSR$\uparrow$ & Params$ \downarrow $ \\
\midrule
Add & 92.5  & 68.5 & 101.8M  \\
Transformer  &  93.3 & 69.0  & 104.8M  \\
Mamba  &  92.7 & 68.1 & 103.5M  \\
\rowcolor[gray]{0.92}
\rule{0pt}{2.0ex}Ours & \textbf{93.8}  & \textbf{69.5} & \textbf{101.8M} \\
\noalign{\hrule height 0.8pt}
\end{tabular}
}
\vspace{-1.1em}
\end{table}
\begin{table}[t]
\centering
\caption{Comparison of token configurations on RGBT234.}
\label{tab:combined}
\vspace{-0.9em}
\newcolumntype{C}[1]{>{\centering\arraybackslash}m{#1}}
\renewcommand\arraystretch{1.0}{
\begin{tabular}{C{0.9cm}C{0.9cm}C{0.9cm}|C{0.9cm}C{0.9cm}C{0.9cm}}
\toprule
\multicolumn{3}{c|}{Reasoning Token} & \multicolumn{3}{c}{Learnable Token} \\
\cline{1-6}
\rule{0pt}{2.25ex}$N_r$ & \rule{0pt}{2.25ex}MPR$\uparrow$ & \rule{0pt}{2.25ex}MSR$\uparrow$ & \rule{0pt}{2.25ex}Length & \rule{0pt}{2.25ex}MPR$\uparrow$ & \rule{0pt}{2.25ex}MSR$\uparrow$ \\
\midrule
\cellcolor[gray]{0.92}1 & \cellcolor[gray]{0.92}\textbf{93.8} & \cellcolor[gray]{0.92}\textbf{69.5} & 0 & 92.8 & 68.4 \\
2 & 93.5 & 69.4 &
\cellcolor[gray]{0.92}2 & \cellcolor[gray]{0.92}\textbf{93.8} & \cellcolor[gray]{0.92}\textbf{69.5} \\
4 & 93.2 & 68.9 & 4 & 92.6 & 68.7 \\
\bottomrule
\end{tabular}
}
\vspace{-1.1em}
\end{table}
\begin{table}[t]
\centering
\caption{Comparison of attention scores on RGBT234.}
\label{tab:attn}
\vspace{-0.9em}
\renewcommand\arraystretch{0.89}{
\begin{tabular}{cccccc}
\toprule
 $\mathbf{A}_{x2r}$ & $\mathbf{A}_{x2h}$ & $\mathbf{A}_{x2z}$ & $\mathbf{A}_{x2x}$ & MPR$\uparrow$  & MSR$\uparrow$ \\
\midrule
  \checkmark & & & &   92.0 &  67.8  \\
  \checkmark & \checkmark & & &   92.9  &  68.7  \\
    \checkmark & \checkmark & \checkmark & &  93.7   & 69.0 \\
  \rowcolor[gray]{0.92}
  \rule{0pt}{2.0ex}\checkmark & \checkmark & \checkmark & \checkmark & \textbf{93.8} & \textbf{69.5} \\
  \noalign{\hrule height 0.8pt}
\end{tabular}
}
\vspace{-1.4em}
\end{table}

\textbf{Effect of Reasoning Token Configurations.} 
We evaluate the effect of reasoning token quantity $N_r$. 
As shown in Tab.~\ref{tab:combined}, the best result is achieved at $N_r=1$, outperforming multi-token configurations. 
This improvement shows that a single reasoning token can capture essential temporal context, while additional tokens may introduce redundant parameters without substantial performance benefits. 
In addition,  our design effectively maintains temporal coherence while minimizing computational overhead.

\textbf{Effect of Learnable Token Lengths.} 
We analyze the effect of learnable token lengths in the sequence prefix. 
As shown in Tab.~\ref{tab:combined}, the 2-token configuration achieves the optimal performance. 
The 0-token variant suffers a 1.0\% MPR drop, highlighting its limited capacity to sequence-specific semantic variations. 
The 4-token variant exhibits a 0.8\% MSR decline, indicating that excessive tokens introduce redundant information. 
These results confirm the necessity of contextual tokens for accurate target representation.

\textbf{Effect of Attention Scores in ATF.} 
We investigate the contribution of different attention scores in ATF. 
As presented in Tab.~\ref{tab:attn}, performance improves progressively as more attention scores are incorporated. 
The complete configuration utilizing all four attention scores achieves the optimal results. 
Our experiments confirm that our multi-score fusion strategy enables effective token selection.
\begin{figure}[t]
  \centering
  \includegraphics[width=0.9\linewidth]{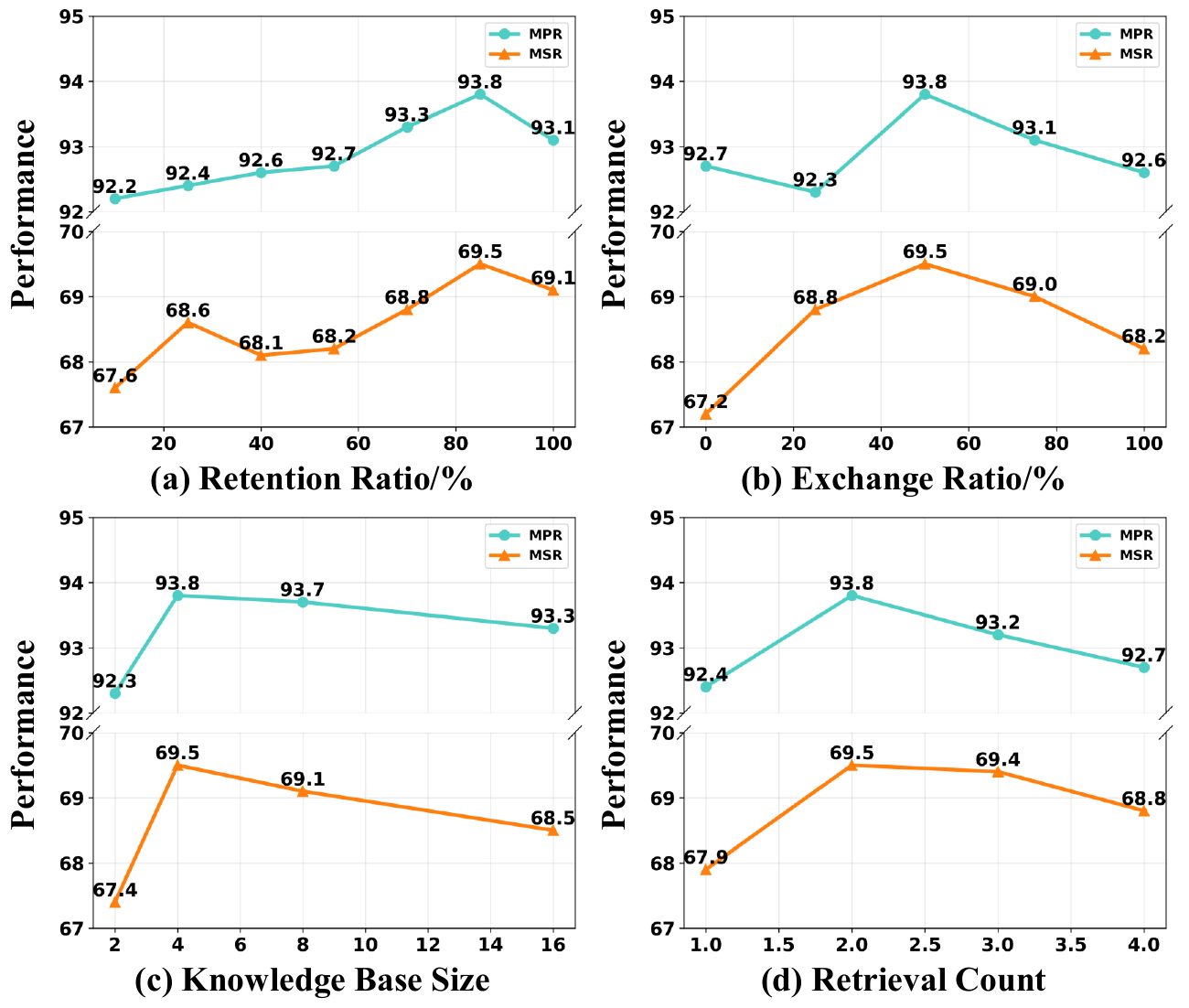}
  \vspace{-3mm}
  \caption{Comparison with different hyper-parameters.}
  \label{fig:impact}
  \vspace{-5mm}
\end{figure}

\textbf{Effect of the Retention Ratio in ATF.} 
The retention ratio determines the proportion of search tokens selected for ATF. 
As shown in Fig.~\ref{fig:impact} (a), $\gamma=85\%$ achieves the leading results with 93.8\% MPR. 
Lower values result in insufficient target information retention, impairing the localization accuracy. 
In contrast, higher values retain noisy or redundant tokens, degrading feature discrimination. 
These findings demonstrate that our setting preserves critical target information while filtering noise.

\textbf{Effect of the Exchange Ratio in ATF.} 
The exchange ratio $\sigma$ plays a key role in cross-modal interaction. 
According to Fig.~\ref{fig:impact} (b), it yields peak performance with $\sigma=50\%$. 
At $\sigma=0$, poor modality fusion results in suboptimal performance. 
As $\sigma$ increases to 50\%, the improved feature representation enhances target modeling. 
However, further increasing $\sigma$ to 100\% introduces excessive cross-modal interference, leading to performance degradation. 
This validates $\sigma=50\%$ as the optimal configuration, achieving a balance between modality specificity and shared representation.

\textbf{Effect of the Knowledge Base Size in CRM.} 
We evaluate how the knowledge base size $n$ affects the tracking accuracy. 
As evidenced by Fig.~\ref{fig:impact} (c), setting $n=4$ achieves 69.5\% MSR. 
When $n=2$, the model lacks sufficient historical context, resulting in a limited reasoning ability. 
While expanding to $n=8$ maintains competitive results, further increasing to $n=16$ causes performance degradation. 
This decline is likely due to the introduction of redundant or less relevant features. 
Thus, $n=4$ strikes the trade-off between context richness and feature quality.

\textbf{Effect of the Retrieval Count in CRM.}
As shown in Fig.~\ref{fig:impact} (d), our method achieves outstanding results with $k=2$. 
Smaller counts compromise performance due to limited contextual information, while larger counts perform worse as excessive features introduce noise. 
By retrieving two relevant features, our method ensures sufficient contextual guidance while achieving effective noise suppression.

\begin{figure}[t]
  \centering
  \includegraphics[width=1.0\linewidth]{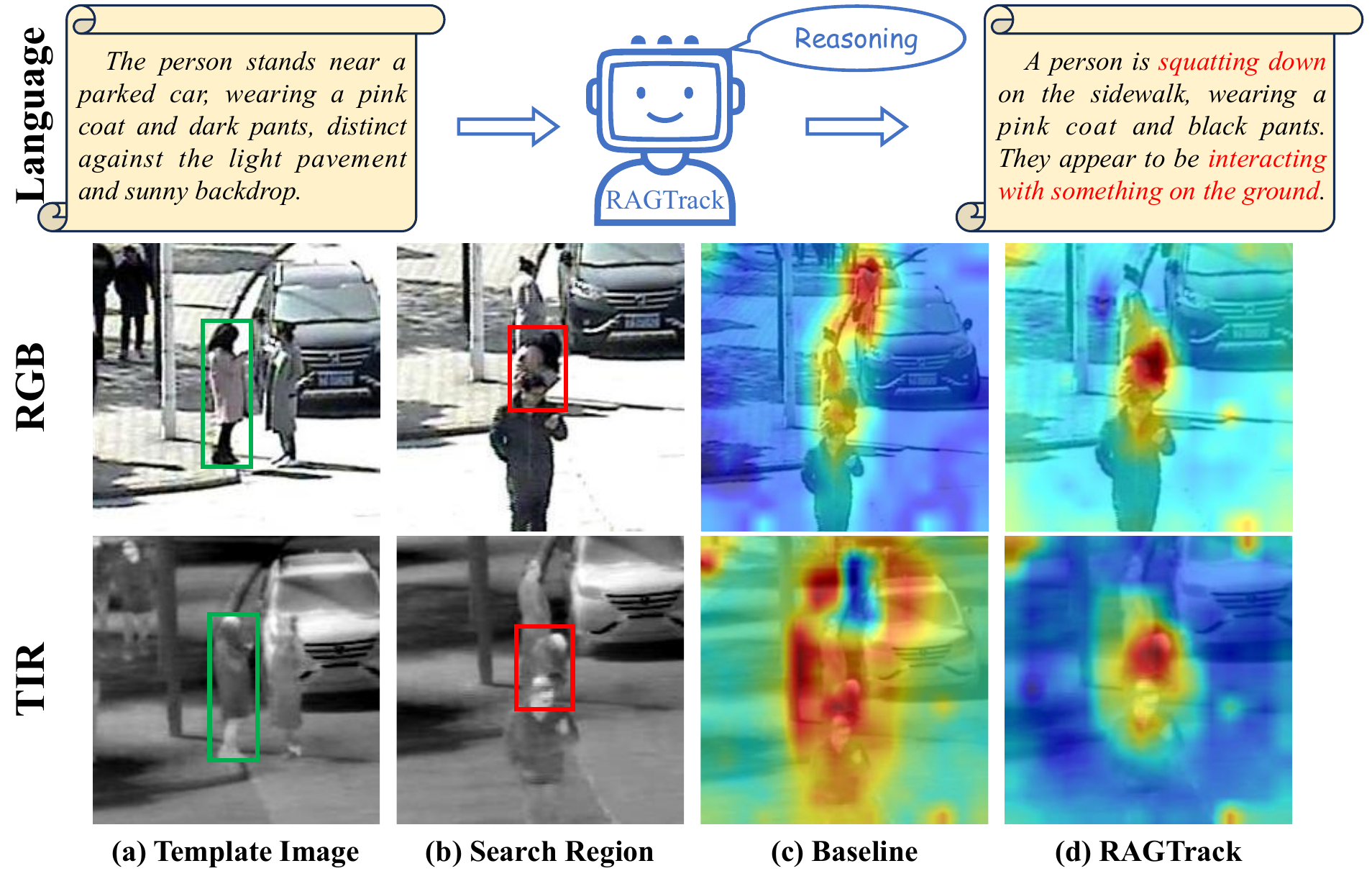}
  \vspace{-5mm}
  \caption{Visualization of attention maps.}
  \label{fig:attention}
  \vspace{-5mm}
\end{figure}

\subsection{Visualization Analysis}
Fig.~\ref{fig:attention} visualizes attention maps across the search region, illustrating how our RAGTrack improves target localization. 
The baseline model shows a noticeable discrepancy between the attentive region and the actual target location. 
In contrast, our proposed method eliminates this misalignment, enhancing the tracking accuracy. 
\section{Conclusion}

In this work, we propose a novel language-aware framework called RAGTrack for RGBT tracking. 
To establish effective feature representation, we introduce a Multi-modal Transformer Encoder (MTE) that aligns visual and linguistic features through unified modeling. 
To reduce search redundancies and modality gaps, we design an Adaptive Token Fusion (ATF), which dynamically selects target-relevant tokens and enables adaptive channel exchange. 
Furthermore, we develop a Context-aware Reasoning Module (CRM) that maintains a dynamic knowledge base and performs context-aware linguistic reasoning to enhance temporal modeling. 
Extensive experiments on four tracking benchmarks validate the effectiveness of our method.

\section*{Acknowledgements}
This work was supported in part by the National Natural Science Foundation of China (No.62441231, 62576069), Dalian Science and Technology Innovation Fund (No.2023JJ11CG001) and Natural Science Foundation of Liaoning Province (No.2025-MS-025, 2023JH26/10200016).
{
    \small
    \bibliographystyle{ieeenat_fullname}
    \bibliography{main}
}
\newpage 
\maketitlesupplementary
This supplementary material provides additional implementation details, experimental analyses, and visualizations for our proposed method. 
We begin by detailing the pipeline for generating high-quality textual descriptions in RGB-Thermal (RGBT) tracking benchmarks (\textsection\hyperref[text generation]{A}). 
We then present the architecture of our prediction head (\textsection\hyperref[prediction head]{B}) and elaborate on the evaluation metrics used in our experiments (\textsection\hyperref[evaluation metrics]{C}). 
Next, we outline implementation details including data augmentation and parameter settings (\textsection\hyperref[implementation details]{D}).
Furthermore, we conduct extensive ablation studies to analyze key components of our method, including fusion position selection and robustness to missing text (\textsection\hyperref[ablation studies]{E}). 
Finally, we show additional visualizations that demonstrate the effectiveness of our dynamic token selection mechanism and present qualitative results under various challenging scenarios (\textsection\hyperref[more visualizations]{F}).
Collectively, these contents offer deeper insights into the design choices of RAGTrack, further validating the overall performance of our framework.

\section*{A. Textual Description Generation}
\label{text generation}
We generate high-quality textual descriptions for RGBT tracking through a two-stage pipeline, ensuring semantic consistency and minimal hallucinations.

\textbf{Step 1: Initial Description Generation.}
We utilize Multimodal Large Language Models (MLLMs) to automatically generate initial descriptions for the object of tracking in each frame. 
The specific prompt used is:
\begin{quote}
\textit{``Describe the object located in the image at $<$box$>$ ({x}, {y}, {x+w}, {y+h}) $<$/box$>$. Focus on distinctive visual features, motion patterns, and key identifiers to distinguish it from background elements and distractors. Keep the description in a continuous sentence under 20 words. Avoid mentioning bounding boxes or coordinates. Do not use parentheses for explanations.''}
\end{quote}

This approach efficiently produces informative descriptions at scale, overcoming the cost of manual annotation.

\textbf{Step 2: Description Refinement.}
The initial descriptions from MLLMs may contain inaccuracies or hallucinations. 
To ensure quality, we perform a refinement step using the following prompt:
\begin{quote}
\textit{``Correcting the textual description of the tracking object. Ensure the final output is a continuous sentence under 20 words, logically coherent, does not mention bounding boxes, or coordinates terms, and does not use parentheses for explanations. Do not introduce new details. Output only the integrated description without any additional text. Textual description: [Initial description]''}
\end{quote}

Finally, human experts review the refined descriptions to correct any remaining issues. 
This includes rectifying hallucinations, fixing grammatical errors, removing mixed-language content or garbled text, and ensuring descriptions accurately reflect the visual target.

\textbf{Annotation Statistics.}
Using this pipeline, we annotate the entire LasHeR training set, which comprises 979 sequences.
This process yields a total of 514,081 textual descriptions, with one description provided for each frame, to support model training. 
For evaluation, we annotate only the first frame across the test set of LasHeR and all sequences of the GTOT, RGBT210 and RGBT234 benchmarks. 
This results in a collection of 739 high-quality textual descriptions, which are used to assess the textual reasoning capability of trackers during inference.

\section*{B. Prediction Head}
\label{prediction head}
The enhanced features of search region are fed into the prediction head to produce tracking results in the form of bounding boxes $\left[ {x,y,w,h} \right]$.
The prediction head generates three outputs: 
(1) a target classification score map $\mathbf{I} \in \left[ {0,1} \right]^{H_F \times W_F}$ indicating presence probabilities, 
(2) a spatial offset map $\mathbf{G} \in \left[ {0,1} \right]^{2 \times H_F \times W_F}$ compensating for discretization errors, 
and (3) a normalized bounding box size map $\mathbf{J} \in \left[ {0,1} \right]^{2 \times H_F \times W_F}$ representing target width and height. 
Here, $H_F$ and $W_F$ denote the height and width of the feature map. 
The final bounding box is constructed at the position $({i^ * },{j^ * })$ with maximum classification score by combining the corresponding predictions:
\begin{equation}
\begin{aligned}
x &= i^* + \mathbf{G}(0,i^*,j^*), & y &= j^* + \mathbf{G}(1,i^*,j^*), \\
w &= \mathbf{J}(0,i^*,j^*),       & h &= \mathbf{J}(1,i^*,j^*).
\end{aligned}
\end{equation}

\section*{C. Evaluation Metrics}
\label{evaluation metrics}
We employ Precision Rate (PR) and Success Rate (SR) as our primary evaluation metrics. 
On the LasHeR benchmark, we additionally use the Normalized Precision Rate (NPR) to address scale variations.
For GTOT and RGBT234, we instead report Maximum Precision Rate (MPR) and Maximum Success Rate (MSR) due to modality annotation misalignment.
PR quantifies the accuracy of target localization as the proportion of frames where the predicted center position lies within a predefined distance threshold from the ground truth. 
SR measures the bounding box overlap, computed as the percentage of frames where the Intersection over Union (IoU) between the predicted and ground truth boxes surpasses a given threshold.
NPR extends PR by normalizing the precision based on target size, providing fair comparison across scale variations.
MPR and MSR address annotation inconsistencies by reporting the maximum performance across modalities, ensuring equitable evaluation when RGB and TIR annotations are misaligned.

\section*{D. Additional Implementation Details}
\label{implementation details}
During training, we apply standard data augmentation techniques~\cite{ke2025early,shan2026hdvs} to the training samples from each sequence, including rotation, translation, and grayscale transformation. 
During inference, the update threshold for multi-modal references is set to 0.65, with an update interval of 5 frames. 
Following common practice~\cite{li2025cadtrack,lu2025rgbt}, the backbone achieves a tracking speed of 24.3 FPS on a NVIDIA V100 GPU, with a computational cost of 62.7G FLOPs.

\section*{E. More Ablation Studies}
\label{ablation studies}

This section presents further ablation studies to examine the impact of individual components in our method. 
Detailed analysis and discussions are provided as follows.

\textbf{Selection of Fusion Positions in ATF.}
To evaluate the impact of fusion locations in Adaptive Token Fusion (ATF), we conduct an ablation study by applying cross-modal fusion at different layers of the Multi-modal Transformer Encoder (MTE). 
As shown in Tab.~\ref{tab:position_ablation}, fusing at Layers 6, 12, 18, and 24 achieves the best performance with 93.8\% MPR and 69.5\% MSR on RGBT234.
Shallow-layer fusion (1-4) captures low-level features but yields suboptimal performance due to limited semantic information.
Mid-layer fusion (11–14) improves semantic understanding but still lacks comprehensive representation.
While deep-layer fusion (21–24) retains high-level context, it misses fine-grained spatial details. 
Our design progressively integrates features across multiple stages, effectively combining spatial details with semantic abstractions. 
The results confirm that cross-layer fusion is essential for robust tracking.

\begin{table}[t]
\centering
\caption{Comparison of different fusion positions.}
\label{tab:position_ablation}
\begin{tabular}{cccc}
\toprule
Fusion Positions & MPR$\uparrow$ & MSR$\uparrow$  \\
\midrule
$\left[ {1,2,3,4} \right]$  & 92.4 & 67.5 \\
$\left[ {11,12,13,14} \right]$  & 92.8 & 67.9 \\
$\left[ {21,22,23,24} \right]$  & 93.1 & 68.7 \\
\rowcolor[gray]{0.92}
$\left[ {6,12,18,24} \right]$ & \textbf{93.8} & \textbf{69.5}  \\
  \noalign{\hrule height 0.8pt}
\end{tabular}
\end{table}

\begin{figure}[t]
  \centering
  \includegraphics[width=1.0\linewidth]{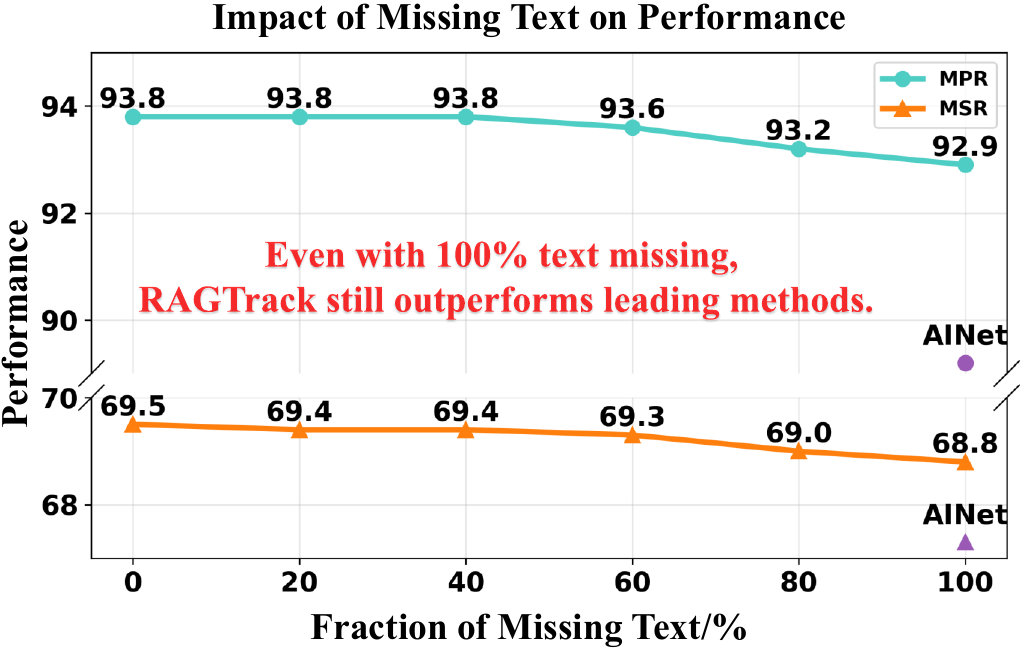}
  \caption{Comparison with different fractions of missing text.}
  \label{fig:text_missing}
\end{figure}

\textbf{Robustness to Missing Text.}
To evaluate the robustness of our method under incomplete language guidance, we conduct an ablation study by randomly masking the input text during inference. 
As shown in the Fig.~\ref{fig:text_missing}, RAGTrack maintains strong performance even when the first-frame text is partially or fully absent. 
The performance remains nearly unchanged with 0\% to 60\% of text missing. 
This demonstrates that our model effectively addresses absent language cues through its Retrieval-Augmented Generation (RAG) mechanism and context-aware reasoning. 
Even when all text is unavailable, our method still achieves competitive results of 92.9\% MPR and 68.8\% MSR on RGBT234, surpassing leading methods~\cite{lu2025rgbt}. 
This highlights the capacity of RAGTrack to robustly utilize visual features when textual input is unavailable.

\begin{figure*}[t]
  \centering
  \includegraphics[width=1.0\textwidth]{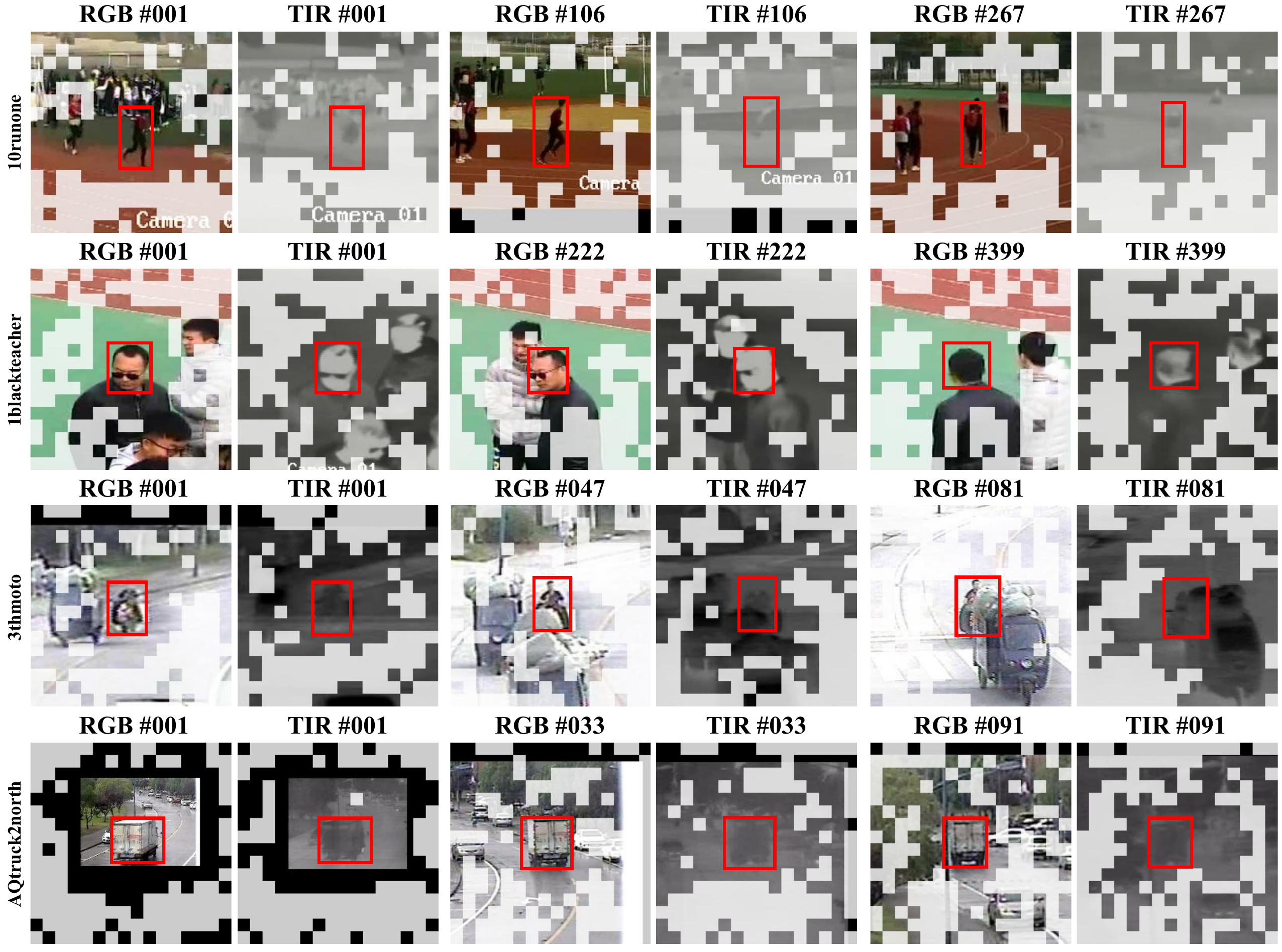}
  \caption{Visualization of dynamic token selection in ATF.
  }
  \label{fig:token_selection} 
\end{figure*}

\section*{F. More Visualizations} 
\label{more visualizations}
This section provides additional visual analysis to further validate the effectiveness of our method through the following examples and discussions.

\textbf{More Discussions of Dynamic Token Selection in ATF.}
To better understand the behavior of our dynamic token selection mechanism in ATF, we provide visualizations of the selected tokens on the LasHeR  benchmark. 
The visualization shows that the mechanism effectively focuses on target regions while suppressing background distractions. 
As shown in Fig.~\ref{fig:token_selection}, the retained tokens mainly cover the target area, while the discarded tokens correspond primarily to background regions and distractors. 
This demonstrates that our attention-based selection identifies semantically important regions guided by the textual descriptions.
Compared to processing all tokens equally, our method reduces unnecessary token processing while maintaining critical target information. 
This selective processing allows the model to concentrate its reasoning capacity on the most informative image regions.
These results provide clear evidence that our dynamic token selection addresses search redundancies, enabling more efficient and accurate RGBT tracking.

\begin{figure*}[t]
  \centering
  \includegraphics[width=1.0\textwidth]{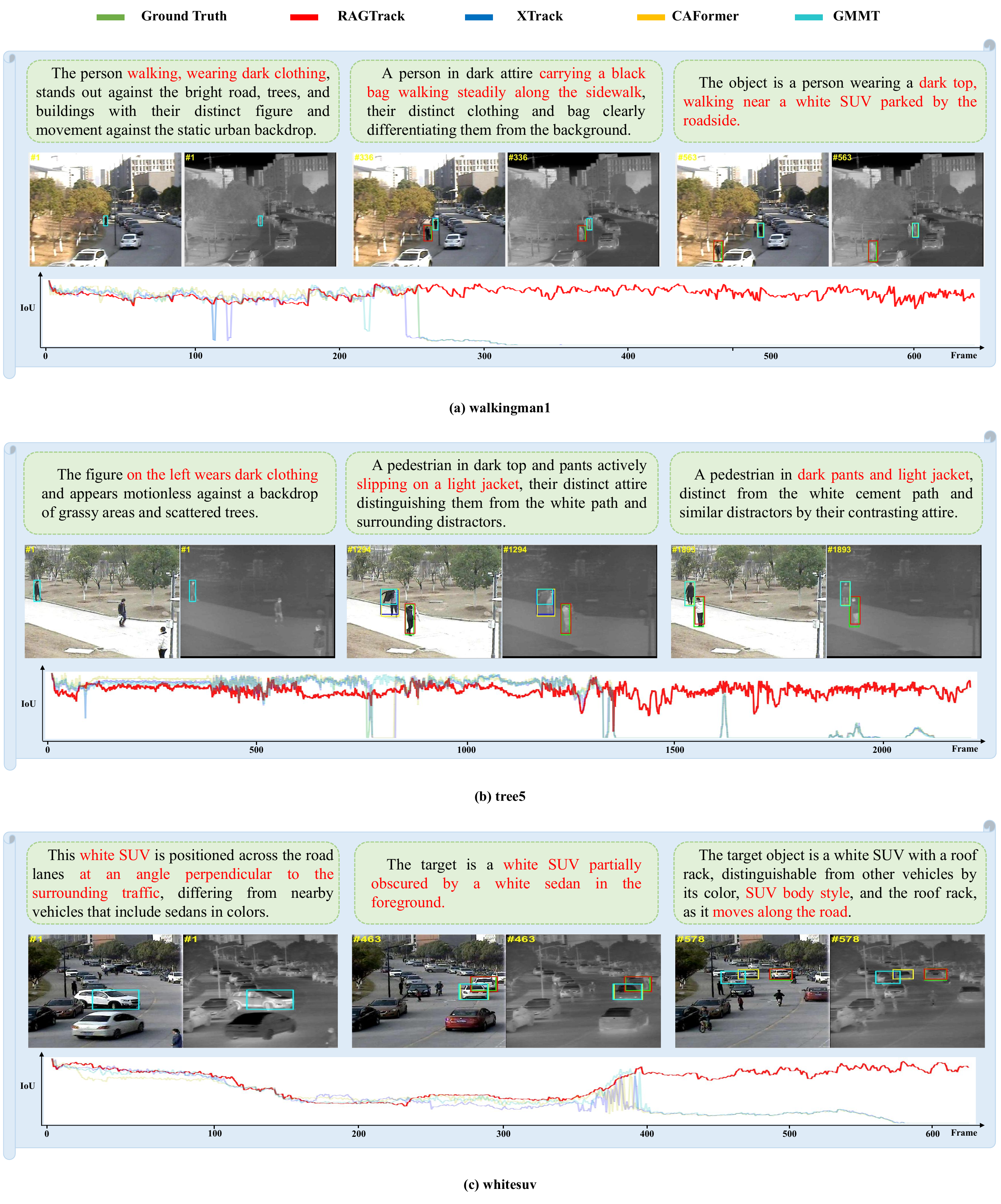}
  \caption{Qualitative results on the RGBT234 benchmark.
  }
  \label{fig:more_tracking_results0} 
\end{figure*}

\begin{figure*}[t]
  \centering
  \includegraphics[width=1.0\textwidth]{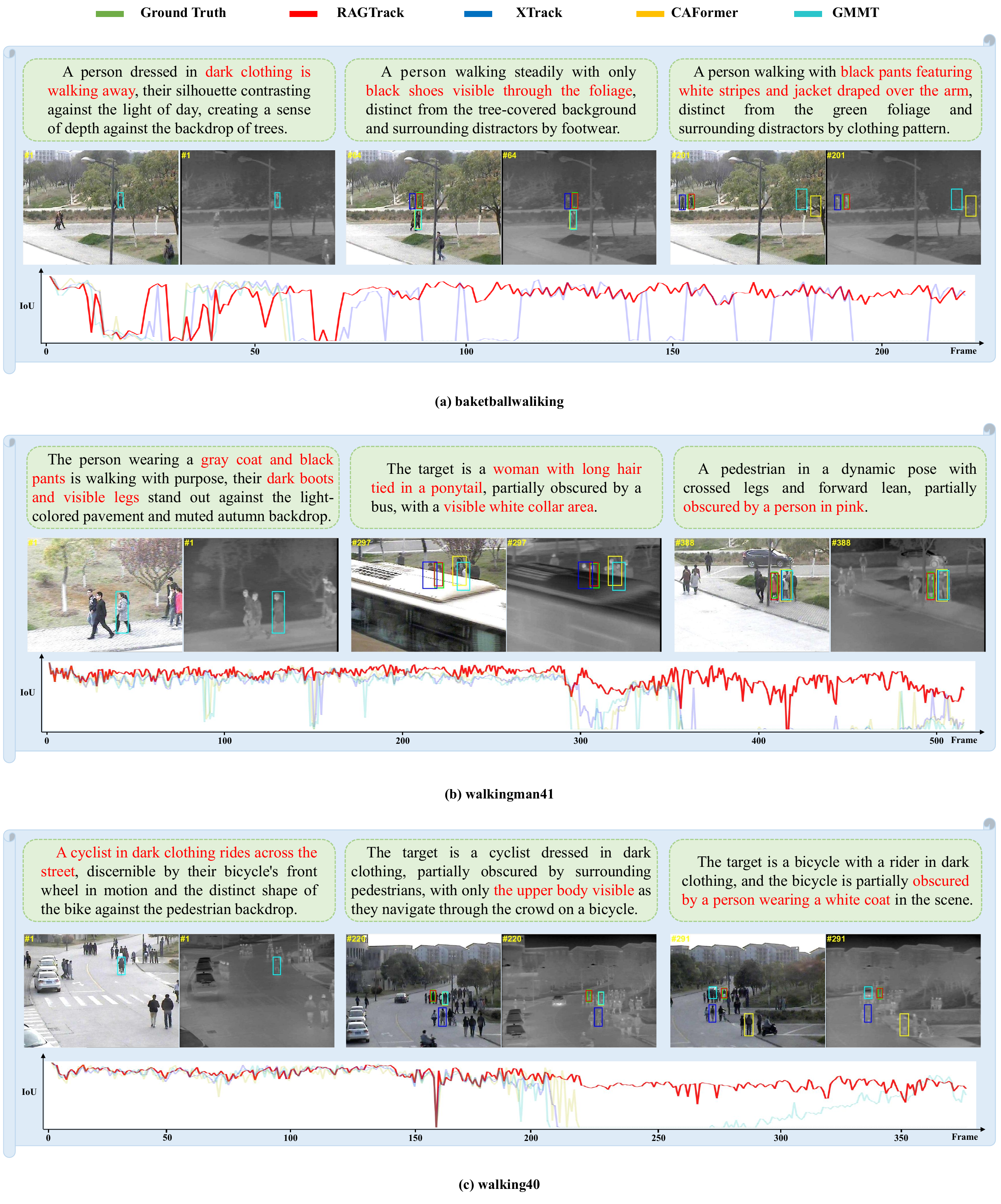}
  \caption{More qualitative results on the RGBT234 benchmark.
  }
  \label{fig:more_tracking_results1} 
\end{figure*}

\textbf{More Qualitative Results.}
Fig.~\ref{fig:more_tracking_results0} and Fig.~\ref{fig:more_tracking_results1} present comprehensive qualitative comparisons of RAGTrack on challenging sequences from the RGBT234 benchmark.
The visualization highlights the capacity of our method to maintain precise tracking through dynamic language reasoning across diverse scenarios. 
Each sequence illustrates the tracking results, evolving textual descriptions and corresponding per-frame IoU curves.
These results demonstrate how our framework effectively handles several challenging situations. 
Through context-aware reasoning and historical knowledge retrieval, RAGTrack successfully distinguishes similar-appearance targets in Fig.~\ref{fig:more_tracking_results0} (a) and Fig.~\ref{fig:more_tracking_results1} (a). 
The method resolves ambiguous target references in Fig.~\ref{fig:more_tracking_results0} (b) by leveraging linguistic guidance to maintain tracking consistency. 
During occlusion shown in Fig.~\ref{fig:more_tracking_results0} (c) and Fig.~\ref{fig:more_tracking_results1} (b), the visual-language unified modeling preserves target identity despite severe appearance changes. 
Additionally, as evidenced in Fig.~\ref{fig:more_tracking_results1} (c), our method overcomes insufficient visual cues through adaptive fusion of complementary multi-modal features.
The stable IoU curves across challenging sequences confirm the robustness of our method in addressing complex tracking difficulties.

\end{document}